\documentclass{article}

\usepackage{microtype}
\usepackage{graphicx}
\usepackage{subfigure}
\usepackage{booktabs} 

\usepackage[accepted]{icml2024}

\usepackage{amsmath}
\usepackage{amssymb}
\usepackage{mathtools}
\usepackage{amsthm}

\usepackage[capitalize,noabbrev]{cleveref}

\theoremstyle{plain}

\theoremstyle{definition}

\theoremstyle{remark}

\usepackage[textsize=tiny]{todonotes}
\usepackage{xcolor}         %
\definecolor{myblue}{rgb}{0.00, 0.45, 0.85}
\definecolor{mygreen}{rgb}{0.05, 0.70, 0.60}
\definecolor{myred}{rgb}{0.84, 0.25, 0.00}
\definecolor{myorange}{rgb}{0.75, 0.25, 0.10}

\icmltitlerunning{Deep Grokking: Would Deep Neural Networks Generalize Better?}

\begin{document}

\twocolumn[
\icmltitle{Deep Grokking: Would Deep Neural Networks Generalize Better?}

\icmlsetsymbol{equal}{*}

\begin{icmlauthorlist}
\icmlauthor{Simin Fan}{epfl}
\icmlauthor{Razvan Pascanu}{deepmind}
\icmlauthor{Martin Jaggi}{epfl}
\end{icmlauthorlist}

\icmlaffiliation{epfl}{EPFL}
\icmlaffiliation{deepmind}{Google Deepmind}

\icmlcorrespondingauthor{Simin Fan}{simin.fan@epfl.ch}
\icmlcorrespondingauthor{Razvan Pascanu}{razp@google.com}
\icmlcorrespondingauthor{Martin Jaggi}{martin.jaggi@epfl.ch}

\icmlkeywords{Grokking, Generalization}

\vskip 0.3in
]

\printAffiliationsAndNotice{}  

\begin{abstract}
Recent research on the \textit{grokking} phenomenon has illuminated the intricacies of neural networks' training dynamics and their generalization behaviors. \textit{Grokking} refers to a sharp rise of the network's generalization accuracy on the test set, which occurs long after an extended overfitting phase, during which the network perfectly fits the training set.
While the existing research primarily focus on shallow networks such as 2-layer MLP and 1-layer Transformer, we explore \emph{grokking} on deep networks (e.g. 12-layer MLP). We empirically replicate the phenomenon and find that deep neural networks can be more susceptible to \emph{grokking} than its shallower counterparts. Meanwhile, we observe an intriguing multi-stage generalization phenomenon when increase the depth of the MLP model where the test accuracy exhibits a secondary surge, which is scarcely seen on shallow models. 

We further uncover compelling correspondences between the decreasing of feature ranks and the phase transition from overfitting to the generalization stage during \emph{grokking}. 
Additionally, we find that the multi-stage generalization phenomenon often aligns with a double-descent\footnote{We distinguish our observation from previously observed double-descent on the loss curves.} pattern in feature ranks. These observations suggest that internal feature rank could serve as a more promising indicator of the model's generalization behavior compared to the weight-norm.

We believe our work is the first one to dive into grokking in deep neural networks, and investigate the relationship of feature rank and generalization performance. 
\end{abstract}

\begin{figure*}[ht!]
    \centering
    \includegraphics[width=\textwidth]{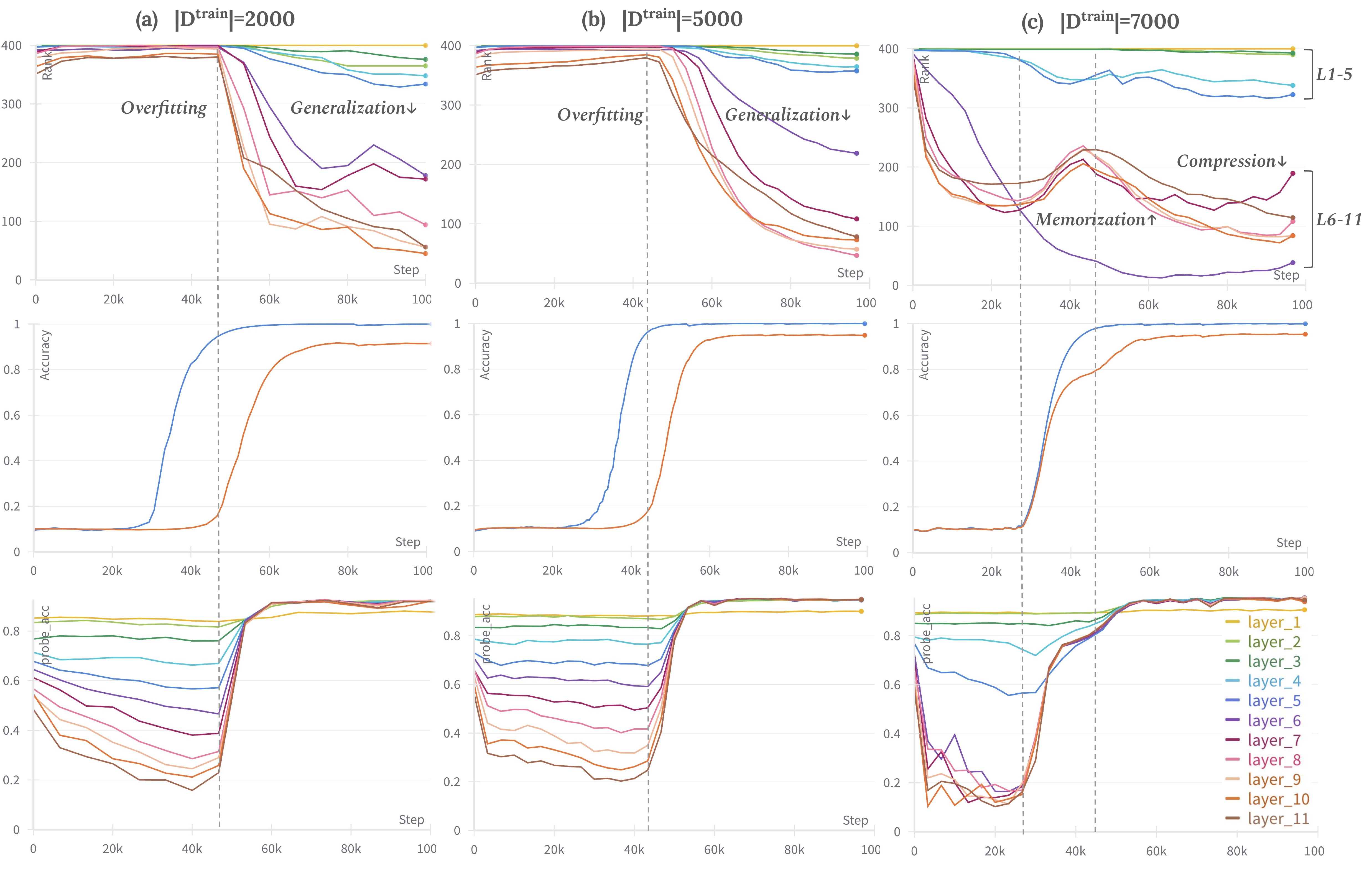}
    \vspace{-2em}
    \caption{\textbf{Generalization behaviors and internal feature learning of a 12-layer MLP network given various amount of training data ($|D^{train}|$).} Middle row depicts training (\textcolor{myblue}{\textbf{blue}} curve) and test accuracy (\textcolor{myorange}{\textbf{orange}} curve) along training steps. Top/Bottom row denotes per-layer feature ranks/linear probing accuracy respectively. \textbf{Left.\&Center. } Grokking-Rank Descent: On small training set, the test accuracy started to increase after the training accuracy reaches the optimal, which corresponds to the feature ranks started to \emph{decrease}. \textbf{Right.} Grokking-Rank Double Descent: On a larger training set, the test accuracy increases in parallel with the training accuracy, which corresponds to an \emph{increment} of feature rank. However, it shows a second surge after training accuracy reaches its optimal, corresponding to a \emph{second descent} of the feature ranks. For all three cases, the increasing of test accuracy always correspond to the increasing of linear probe accuracy across all layers. Different colors of feature ranks represent different layers.} 
    \label{fig:grok-rank}
\end{figure*}
\section{Introduction}
The efficacy of a machine learning algorithm hinges on minimizing the generalization gap: only with a negligible disparity between training and test errors, the empirically optimal solution obtained from the training set could be extended to unseen test sets. 

However, \citet{power2022grokking} has recently observed an unexpected phenomenon named ``grokking" on a simple modular addition task, where the generalization delays after an extended overfitting stage: The test accuracy shows a sharp rise from low to near perfect level long after the training accuracy reach its optimum (Fig. \ref{fig:grok-rank}, (a,b)). 

The community has dedicated substantial efforts in understanding this delayed and sharp phase transition from overfitting to the generalization regime \citep{power2022grokking,liu2022understanding,thilak2022slingshot,liu2023omnigrok,lyu2023dichotomy}, while nearly all existing studies on grokking are confined to \emph{shallow} networks (e.g. 2-layer transformer, 2-layer MLP). However, as evidenced by recent findings, markedly different generalization behaviors can emerge when the model is scaled up \citep{wei2022emergent}. To bridge the gap, we conduct the first inquiry of grokking on deep neural networks (e.g. 12-layer MLPs) and investigate the correlation between evolution of internal feature representations and phase transitions during the training process. 

Following \citet{liu2023omnigrok}, we replicate the grokking phenomenon on deep MLP networks applying a large initialization and small weight decay. Specifically, we presents three intriguing observations on deep MLP networks: 
\begin{enumerate}
    \item Deep MLP networks are more susceptible to grokking than  shallow models;
    \item On deep MLP networks, we observed a correspondence between feature rank decrease and generalization. It suggests that the internal feature ranks could be a more promising indicator of phase transition during training compared to the weight-norm proposed by \citep{liu2023omnigrok};
    \item We observe a multi-stage progress in generalization, where the test accuracy evolves in two sharp increments instead of a single surge observed previously. The two phases are mirrored by a \emph{double-descent} pattern on the feature rank. 
\end{enumerate}

\section{Experiments}
\paragraph{Training Setup.}
Following \citet{liu2023omnigrok}, we study grokking on Multi-Layer Perception (MLP) networks of varying depth on the MNIST dataset \citet{LeCun1998mnist}. We apply large initialization and small weight decay to reproduce grokking, mimicking the set-up of \citet{liu2023omnigrok}. All the models have the same width (400) and implemented with ReLU activation. 
For each network, we scale the standard initialization\footnote{as default initialization in Pytorch.} by a scaling factor $\alpha$.\footnote{$\alpha=w/w_0$, where $w_0$ and $w$ are the weight-norm of the network before and after rescaling.} All the models are trained with the Adam optimizer (learning rate $1\times10^{-3}$) and Mean Square Error (MSE) for $10^5$ steps following the setting in \citet{liu2023omnigrok}. If not otherwise specified, the model is trained with initialization scale $\alpha=8.0$ and weight decay $\gamma=0.01$.

\paragraph{Probing Internal Features.}
We study the internal feature representation learning by measuring the (i) Linear probing accuracy, which indicates the quality of the internal representation and (ii) Numerical rank of representations, which indicates the extent of compression. 
For linear probing, we attach a randomly initialized linear classifier head to each internal layer $l$ of the neural network. We train the linear head only for a fixed number of steps on the training set, then evaluate it on the test set. The linear probing accuracy measures whether the learnt internal features are good enough to be linearly separable in the high-dimensional space. 
We also estimate the layer-wise numerical rank of feature representations to reflect to what extent the learnt internal features can be compressed. Specifically, we compute the singular values of the covariance matrix of the output activation $W\in \mathrm{R}^{n\times d}$ from an internal layer $l$. We then measure its numerical rank as the number of singular values above a certain threshold given a relative tolerance $\epsilon$: $\sigma=max(n,d)\cdot \epsilon$.\footnote{We use as tolerance $\epsilon$ the epsilon value of the dtype of the feature matrix $W$ used Pytorch \textit{matrix\_rank} function.}

\begin{figure*}[ht!]
    \centering
    \includegraphics[width=\textwidth]{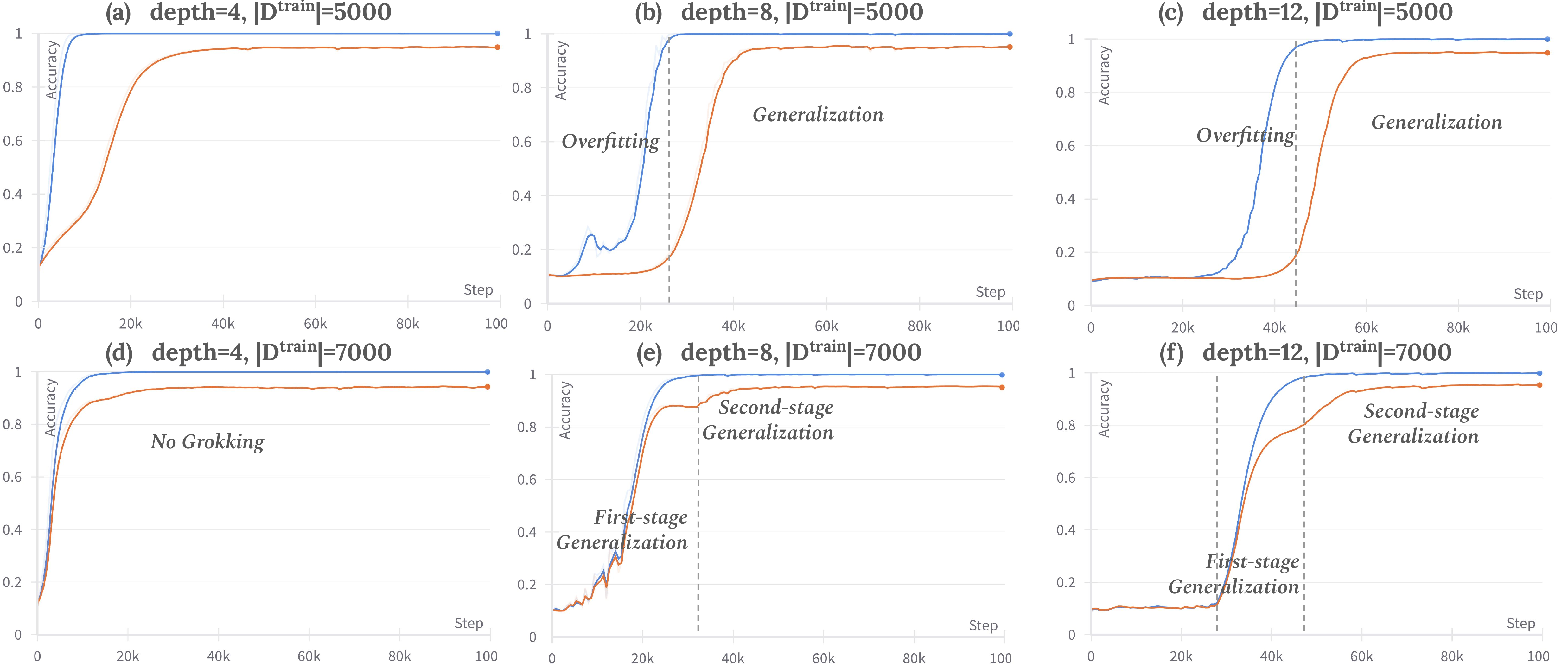}
    \vspace{-1em}
    \caption{\textbf{Generalization behaviors of various depth of MLP models.} \textbf{Top.} Grokking: On small training set ($|D^{train}|=5000$), both the improvement of training (\textcolor{myblue}{\textbf{blue}} curve) and test (\textcolor{myorange}{\textbf{orange}} curve) accuracy would be delayed as the depth of the model increase. \textbf{Bottom.} Multi-stage Generalization: On a larger training set ($|D^{train}|=7000$), the deep (8- and 12-layer) MLP models exhibit a 2-stage generalization, where the test accuracy experiences a second surge. The 2-stage generalization phenomenon is also correlated to a double-descent of feature rank (Fig. \ref{fig:grok-rank}, c).} 
    \label{fig:deep-grok}
\end{figure*}
\subsection{Generalization Behavior of Deep Networks}
\paragraph{Deep networks are more likely to perform \textit{grokking}.} 
The long-standing intuition from deep learning suggests that deeper neural networks have the potential to overfit on the training set as they are more expressive~\cite{montúfar2014number}. They are also meant to learn complex features and are known to generalize better than shallow networks~\cite{zeiler2013visualizing,ozair2014deep}. 
Our experiments reveals that deeper MLP networks are more prone to experiencing \textit{grokking} compared to their shallower counterparts, which implies more sophisticated features learnt by deeper layers not help much with generalization. As illustrated in Figure \ref{fig:deep-grok} (a,b,c), as the network depth increases, both the growth in training and test accuracy are notably delayed.

\paragraph{Multi-stage generalization.}
We also observed an intriguing multi-stage generalization behavior on deep MLP networks, wherein the test accuracy demonstrated several distinct periods of improvement marked by sharp transitions. According to Figure \ref{fig:deep-grok} (d,e,f), when trained on 7000 data points, no significant \textit{grokking} was observed during the training of the 4-layer MLP network, while both the 8-layer and 12-layer MLP networks exhibited a notable second-surge of test accuracy. Meanwhile, compared to the 8-layer, the occurrences of both the first- and second- stage of generalization are delayed with the 12-layer model, and the test accuracy obtained from the first-stage generalization is significantly lower than the 8-layer network. 

Moreover, the multi-stage generalization behavior is corresponding to a double-descent of feature ranks. As depicted in Figure \ref{fig:grok-rank} (c), the feature ranks drop from the beginning of training, while starting to increase at the first surge of test accuracy. Concurrent to the second stage of generalization, the feature ranks experience a second-descent.

\subsection{Internal Feature Learning of Deep Networks}
\paragraph{Correlation between Phase Transitions and Rank Descent.}
We have observed a concurrent feature rank descent corresponding to the phase transition of ``grokking" phenomenon on 12-layer MLP networks. According to Figure \ref{fig:grok-rank} (a), during the training of a 12-layer MLP network on limited data, the test accuracy (and linear probe accuracy) begins to rise notably as the feature ranks exhibit a significant decrease. With larger training data size (Fig. \ref{fig:grok-rank}, b), the process of grokking occurs more rapidly, in accordance with findings by \citet{power2022grokking} and \citet{liu2023omnigrok}. In these instances as well the correlation between rank collapse and the shift from overfitting to generalization stage remains consistent. It
suggests that the feature ranks could be a more
promising indicator of phase transition during training compared to the weight-norm, which is investigated by \citep{liu2023omnigrok}.

\begin{figure*}[ht!]
    \centering
    \includegraphics[width=\textwidth]{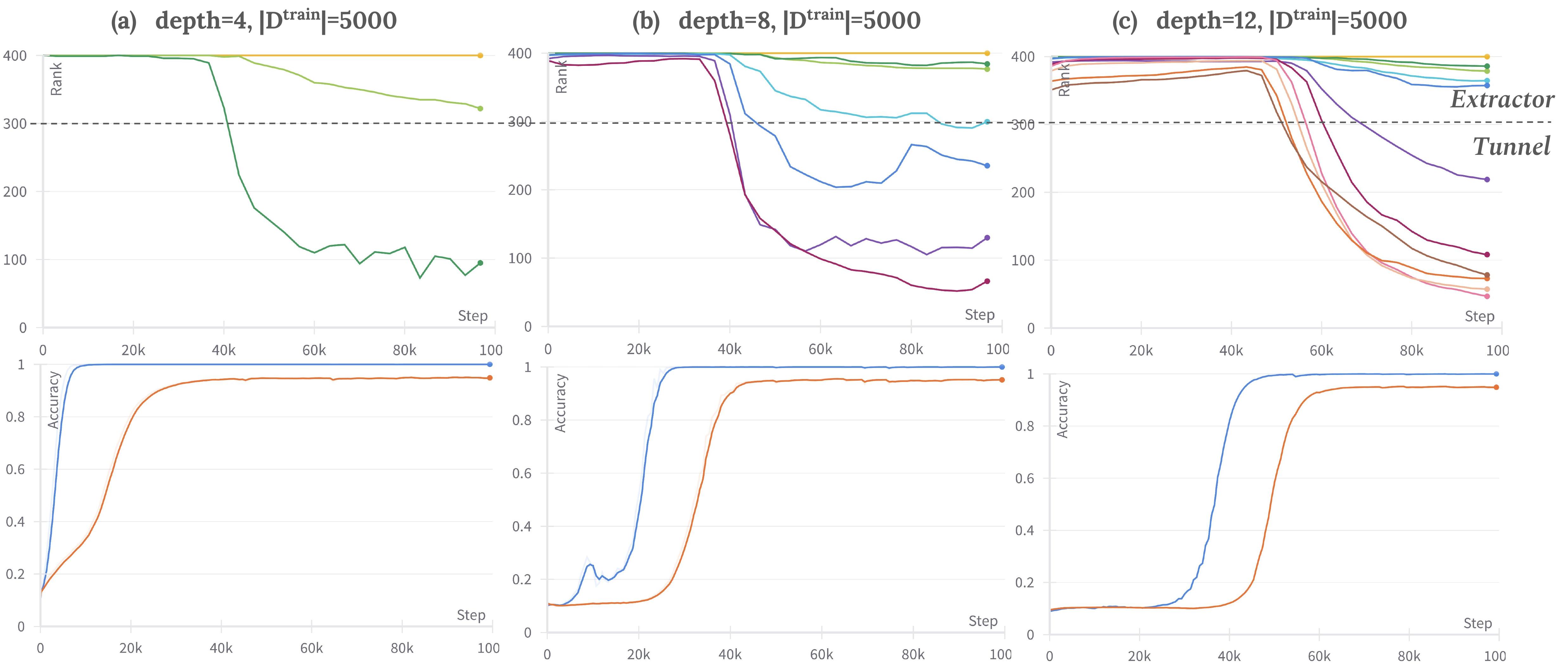}
    \vspace{-1em}
    \caption{\textbf{Emergence of \textit{Tunnel} on various depth of models ($|D^{train}|=5000$).} \textbf{Left.} 4-layer MLP: least extent of \textit{grokking} with $tunnel\_length=1 layer$; \textbf{Middle.} 8-layer MLP: more \textit{grokking} than 4-layer model with $tunnel\_length=3 layer$; \textbf{Right.} 12-layer MLP: the most severe \textit{grokking} with $tunnel\_length=6 layer$.
    We consider the layer with output feature rank lower than 300 as a \textit{tunnel} layer.} 
    \label{fig:tunnel-5000}
\end{figure*}

\begin{figure*}[ht!]
    \centering
    \includegraphics[width=\textwidth]{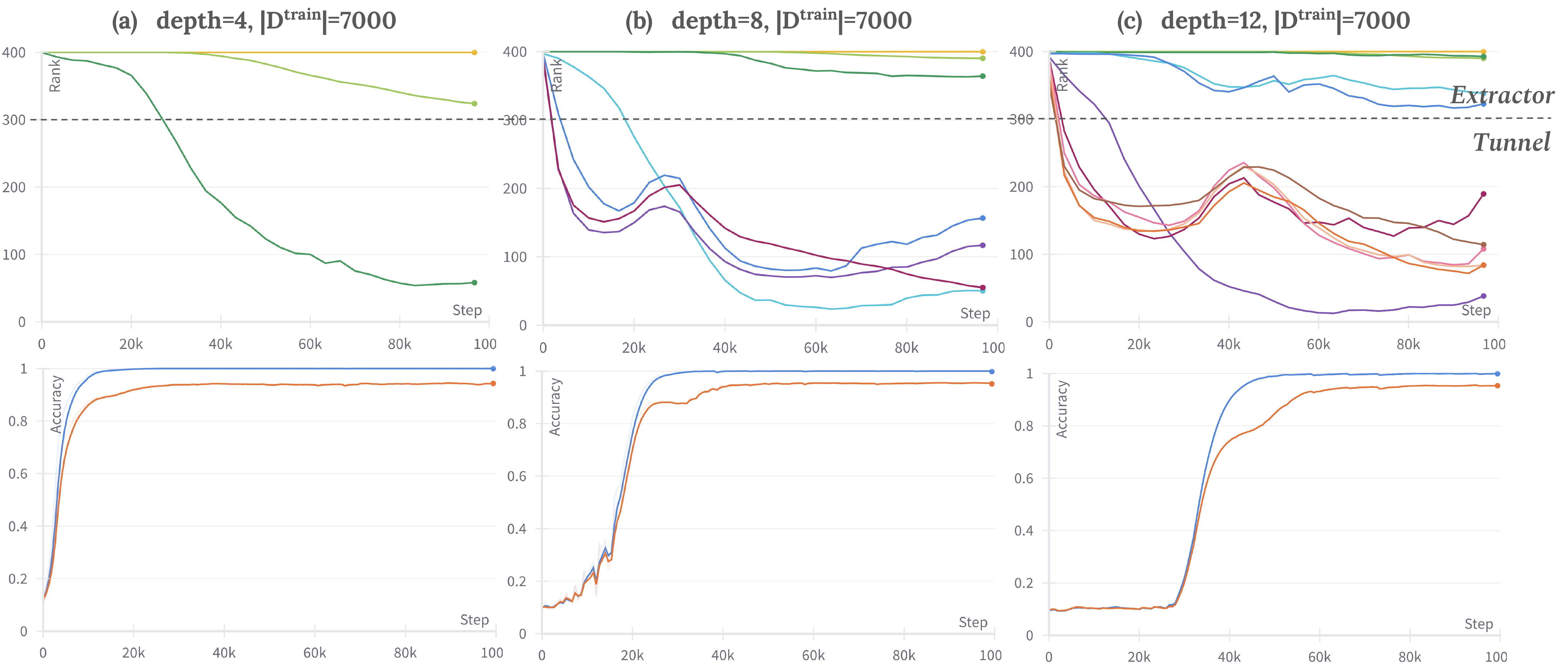}
    \vspace{-1em}
    \caption{\textbf{Emergence of \textit{Tunnel} on various depth of models ($|D^{train}|=7000$).} \textbf{Left.} 4-layer MLP: grearly generalize with no \textit{grokking} with $tunnel\_length=1 layer$; \textbf{Middle.} 8-layer MLP: \textit{two-stage generalization} with $tunnel\_length=5 layer$; \textbf{Right.} 12-layer MLP: \textit{two-stage generalization} with delays in both generalization stages, $tunnel\_length=6 layer$.
    We consider the layer with output feature rank lower than 300 as a \textit{tunnel} layer.} 
    \label{fig:tunnel-7000}
\end{figure*}

\paragraph{Emergence of the "\textit{Tunnel}".}
\citet{masarczyk2023tunnel} demonstrated that the architecture of an image classification model can split into two distinct parts: the initial layers (i.e. \textit{Extractor}) learn a high-rank, linearly separable feature representation while the higher layers (i.e. \textit{Tunnel}) tend to compress it into a low-ranked representation. As exhibited in Fig. \ref{fig:tunnel-5000} and Fig. \ref{fig:tunnel-7000}, our observations align with these findings in two aspects: firstly, as the layer depth increases, the feature ranks decrease more significantly during training; secondly, deeper networks tend to have longer ``\textit{tunnel}"s. The empirical results from \citep{masarczyk2023tunnel} suggest that the longer tunnel may have a detrimental effect on the learned representation's generalization ability. We hypothesize that the similar ``\textit{tunnel effect}" could lead to the more pronounced \textit{grokking phenomenon} within deeper MLP networks compared to the shallower counterpart. 

\subsection{Where is the \textit{Godilocks}? Rethinking the Relationship between Weight-norm and Grokking}
\citet{liu2023omnigrok} presents a hypothesis of \textit{grokking} from large initialization based on the \textit{Godilocks Zone} theory: there exist a thick, hollow,
spherical shell in the space of model's weight around the initialization which could lead to a great generalization behavior \citep{fort2018goldilocks}. According to \citep{liu2023omnigrok}, the large, bad initialization would put the model outside the shell of the \textit{Godilocks zone}, where the model would quickly fit to the training set with a slight change of the weight-norm, while being slowly drifted to the \textit{Godilocks zone} by the weight decay regularization, corresponding to a drop of the weight-norm. However, could the weight-norm be considered as a proper indicator of \textit{grokking}?

\begin{figure*}[ht!]
    \centering
    \includegraphics[width=0.8\textwidth]{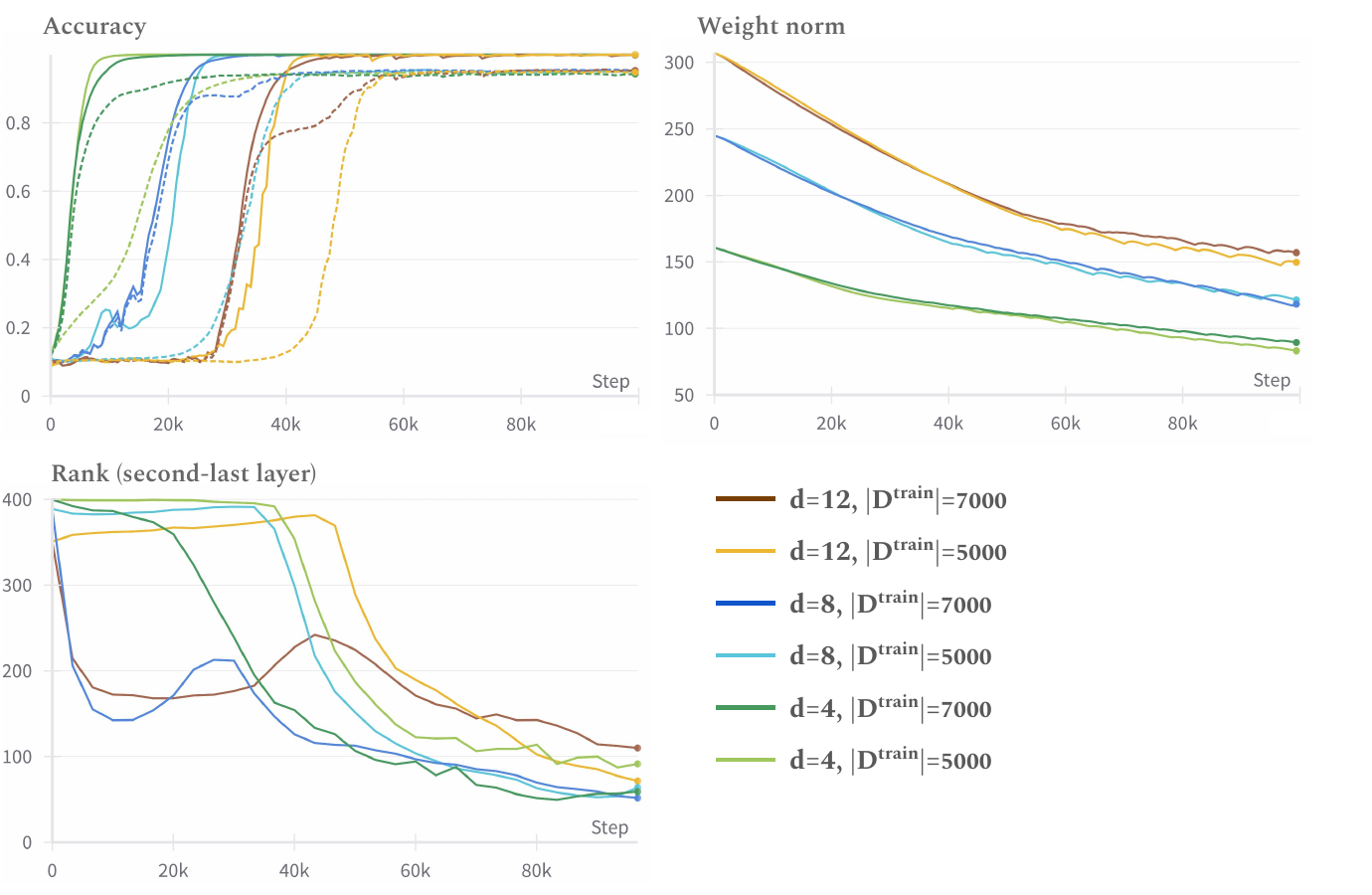}
    \vspace{-1em}
    \caption{\textbf{Trajectories of train/test accuracy, weight-norm and feature-rank.} \textbf{Top-left}: the train and test accuracies. The dotted lines stand for test accuracies; \textbf{Top-right}: the trajectories of the weight-norms, defined as the magnitude of L2-norm of all the model parameters. \textbf{Bottom}: we present the feature rank curve from the \emph{second-last layer} (the layer before output layer). When trained on various scales of training data, the trajectories of the weight-norms are largely overlapping while the feature-ranks exhibit significant differences indicating the phase transitions. The finding holds across various depths of models ($d=4,8,12$).} 
    \label{fig:norm}
\end{figure*}

We observed that \textbf{\emph{the dynamics of the weight-norm are not indicative of the model's generalization behavior.}}
Following the setting as in Figure \ref{fig:deep-grok}, we measure the weight-norm, defined as the L2 norm of all model parameters, in Figure \ref{fig:norm}. 
Specifically, we train two MLP models of the same depth on two different scales of training sets then compare the dynamics of the weight-norm and the feature-rank from the second-last layer during. Despite the distinguishable generalization behaviors, the weight-norm's trajectories are largely overlap, which decrease smoothly throughout the training process, without a discernible indication of the phase transitions. In contrast, the feature-ranks exhibit significant differences indicating the phase transitions. We consolidate our finding with experiments on various depth of models Fig. \ref{fig:norm} ($d=4,8,12$). 

\section{Dependence on Weight Decay}
\begin{figure*}[ht!]
    \centering
    \includegraphics[width=\textwidth]{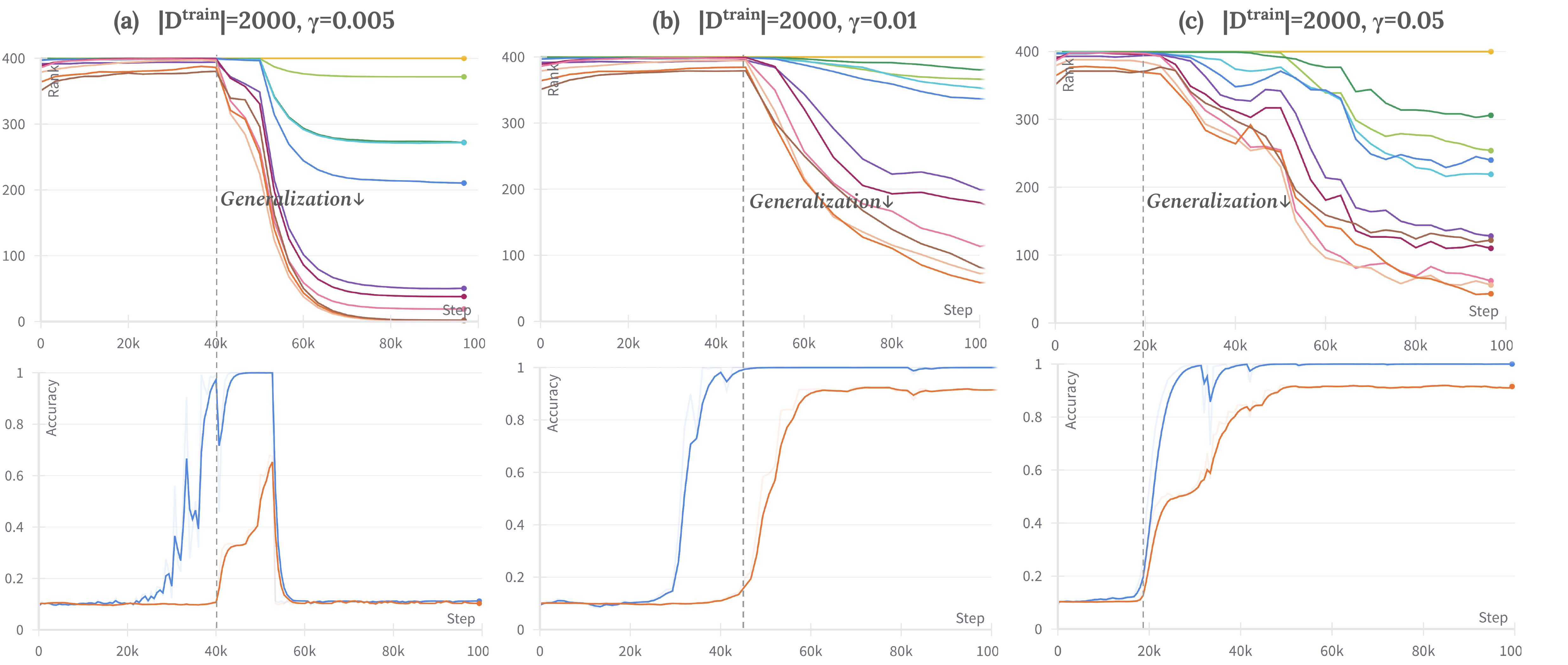}
    \vspace{-1em}
    \caption{\textbf{Generalization behaviors with various weight decay on $|D^{train}|=2000$.} \textbf{Left.} $\gamma=0.005$: Fail to Generalize; \textbf{Middle.} $\gamma=0.01$: Grokking; \textbf{Right.} $\gamma=0.05$: Two-stage Generalization.} 
    \label{fig:wd-2000}
\end{figure*}

Following \citet{liu2023omnigrok}, we also investigate the dependency on the strength of regularization by experimenting with three weight decay values: $\gamma=0.005$, $\gamma=0.01$, $\gamma=0.05$. As shown in Fig. \ref{fig:wd-2000}, on 2000 training data points, the smallest regularization ($\gamma=0.005$) fails to generalize; with $\gamma=0.01$, the model demonstrates a delayed generalization (\textit{grokking}); with the largest regularization ($\gamma=0.05$), the model exhibits a two-stage generalization. With a larger training data size (5000), the model shows similar grokking (resp. two-stage generalization) behaviors with $\gamma=0.01$(resp. $0.05$); while with the smallest regularization ($\gamma=0.005$), the model is able to generalize without grokking. All the training data size and weight decay combinations exhibit a consistent correlation between the beginning of generalization and the descent of the feature ranks. Also, the two-stage generalization phenomenons are aligned with a double-descent of the feature ranks.

\begin{figure*}[ht!]
    \centering
    \includegraphics[width=\textwidth]{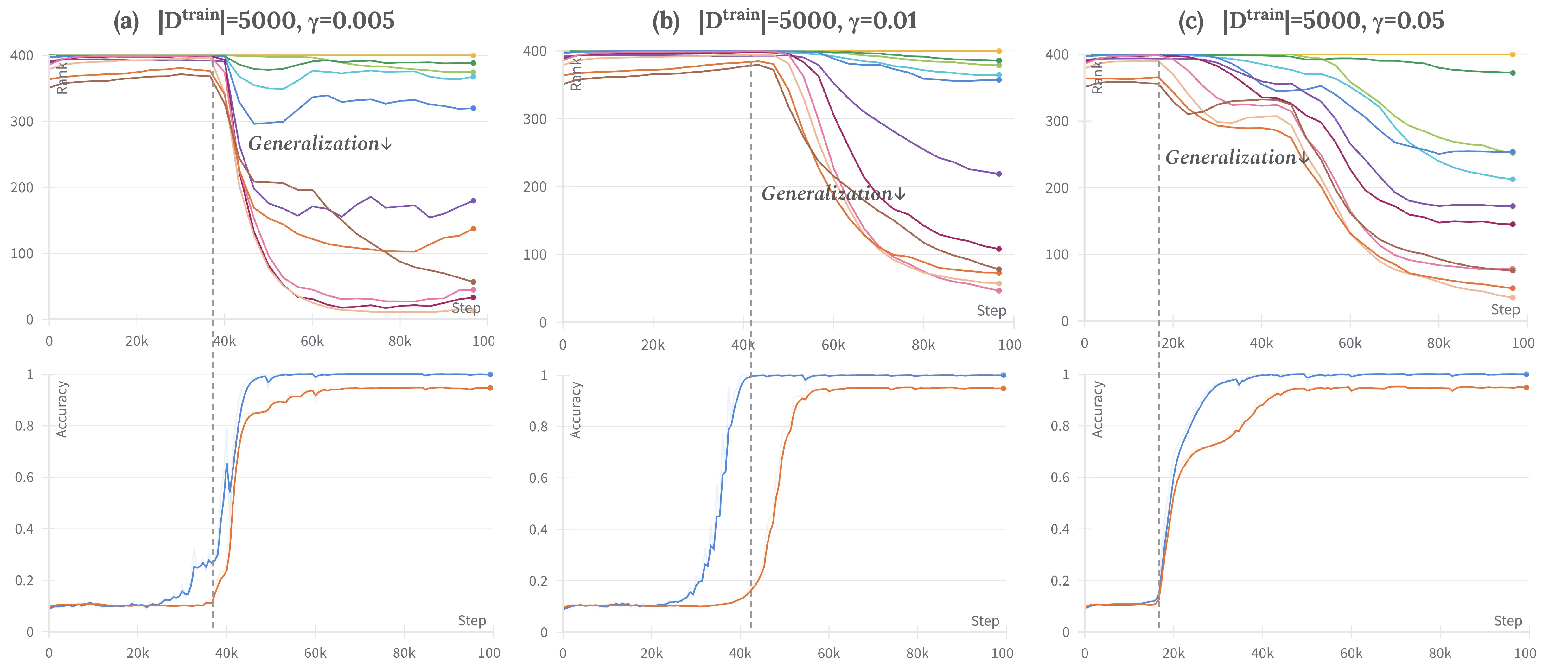}
    \vspace{-1em}
    \caption{\textbf{Generalization behaviors with various weight decay on $|D^{train}|=5000$.} \textbf{Left.} $\gamma=0.005$: Generalize without grokking; \textbf{Middle.} $\gamma=0.01$: Grokking; \textbf{Right.} $\gamma=0.05$: Two-stage Generalization.} 
    \label{fig:wd-5000}
\end{figure*}

\section{Related Work}

\paragraph{Grokking on Shallow Networks.}
\citet{power2022grokking} has first observed the grokking phenomenon on small algorithmic datasets, where the validation accuracy of a 2-layer transformer model performs a sudden increment from chance level toward a perfect generalization long after a severe overfitting stage. Following Power et al., \citet{liu2023omnigrok} suggested that one could induce grokking on various model architectures (MLP and RNN) and tasks by applying large initialization scale and small but non-zero weight decay. \citet{lyu2023dichotomy} further produced a \textit{misgrokking} phenomenon with diagonal linear networks, where the model first well generalizes to 100\% test accuracy, and then drops to nearly 50\% after training for sufficiently longer.

\paragraph{Explanation of Grokking.}
Current research have also provided valuable insights into the \textit{grokking} phenomenon from the dynamics of learning. \citet{thilak2022slingshot} reported that the \textit{grokking} happens exclusively in accordance to a \textit{slingshot effect}, which corresponding to a cyclic phase transitions between stable-and-unstable training regimes and the norm growth-and-plateau of the last layer weights. However, for the multi-stage generalization on deep networks, we have not observed any significant plateaus of the norm of the last layer or the entire set of model parameters. \citet{liu2022understanding} first attributed \textit{grokking} to the slow formulation of good representations. However, they mainly construct a theory between the representation learning effectiveness and the size of available training data, without diving deep into the detailed feature characteristics (i.e. ranks collapse etc.). Following that, \citet{liu2023omnigrok} further explores the correspondence between the weight-norm and the transition of overfitting-generalization regimes, explaining that the model needs to \textit{grok} slowly into the \textit{Goldilocks zone} \citep{fort2018goldilocks} for generalization. \citet{lyu2023dichotomy} explains grokking from the distinct implicit biases induced from overfitting and generalization phases from the \textit{kernel-regime} where the classifier behaves like a Neural Tangent Kernel (NTK) to a \textit{rich-regime}, which mainly based on a gradient-based convergence of a min-max minimizer. 

Concurrently, contemporary studies in mechanistic interpretability provided valuable insights of understanding model training dynamics through the lens of feature ranks \cite{feng2022rank,boixadsera2023transformers,wang2024understanding}. However, none have yet linked layerwise feature ranks to the concept of \textit{grokking}. \citet{andriushchenko2023sharpnessaware} suggests that sharpness-aware minimization would find solutions with lower feature ranks, which echos to our findings of the correlation between rank decrease and generalization.
 Therefore, our research presents the first evidence on the relationship between feature rank and \textit{grokking}, aiming to enhance comprehension of generalization behavior through the analysis of internal feature representations. There are plentiful explorations on the double-descent phenomenon in over-parametrized networks \citep{nakkiran2019deep,kuzborskij2021role,schaeffer2023double,lafon2024understanding}, which could be connected to our observations.
 
\section{Conclusion}
In this paper, we do the first inquiry on the grokking phenomenon and generalization behaviors on deep MLP networks. Unconventionally, we found that deep MLP models are more susceptible to suffer from \textit{grokking} than its shallower counterparts. Also, we have observed the intriguing multi-stage generalization and a consistent feature rank-generalization correspondence on deep MLP models. Our observation suggests that internal feature ranks could serve as a promising indicator of the model's generalization performance. We leave the theoretical analysis and extensive experiments on other architectures (e.g. Transformers, RNNs) to the future work.

\newpage
\bibliography{main}
\bibliographystyle{icml2024}


\end{document}